\documentclass{article}
\usepackage{spconfa4,amsmath,graphicx}
\usepackage{multicol,multirow,booktabs}
\usepackage[citecolor=black,
             colorlinks=true,
             linkcolor=black,
             urlcolor  = black,
             pdftitle={},
             pdfauthor={},
             pdfkeywords={},
             pdfsubject={}]{hyperref}
\usepackage{xcolor}
\usepackage{acronym}
\usepackage{amssymb}

\usepackage[english]{babel}
\addto\extrasenglish{}
\addto\extrasenglish{}
\addto\extrasenglish{}
\addto\extrasenglish{}
\addto\extrasenglish{}
\addto\extrasenglish{}

\title{CHiME-9 ECHI: A Machine Learning Challenge for Enhancing Conversations to Address Hearing Impairment}
\name{Robert Sutherland$^1$\thanks{This work was supported by the UKRI AI Centre for Doctoral Training in Speech and Language Technologies (SLT) and their Applications funded by UK Research and Innovation [grant number EP/S023062/1]. For the purpose of open access, the author has applied a Creative Commons Attribution (CC BY) licence to any Author Accepted Manuscript version arising. This work was also supported by WS Audiology and Meta.}, Thomas Kuebert$^2$, Marko Lugger$^2$, Stefan Petrausch$^2$, Eline Borch Petersen$^3$, Juan Azcarreta Ortiz$^4$, Buye Xu$^5$, Stefan Goetze$^{1,6}$, Jon Barker$^1$}
\address{$^1$School of Computer Science, University of Sheffield, Sheffield, United Kingdom\\
            $^2$WS Audiology, Erlangen, Germany\\
            $^3$ORCA Labs, WS Audiology, Lynge, Denmark\\
            $^4$Meta Reality Labs, Cambridge, United Kingdom\\
            $^5$Meta Reality Labs, Redmond, United States of America\\
            $^6$South Westphalia University of Applied Sciences, Iserlohn, Germany}

\acrodef{AHD}       {assistive hearing device}
\acrodef{CT}        {close-talk}
\acrodef{ECHI}      {Enhancing Conversations to address Hearing Impairment}
\acrodef{fwSegSNR}  {frequency-weighted segmental SNR}
\acrodef{HA}        {hearing aid}
\acrodef{LSTM}      {long-short term memory}
\acrodef{MSX}       {multi-speaker extraction}
\acrodef{NN}        {neural network}
\acrodef{OVS}       {own-voice suppression}
\acrodef{PESQ}      {perceptual evaluation of speech quality}
\acrodef{SNR}       {signal-to-noise ratio}
\acrodef{STFT}      {short-time Fourier transform}
\acrodef{STOI}      {short-time objective intelligibility}
\acrodef{TSX}       {target speaker extraction}
\acrodef{VAD}       {voice activity detection}
\begin{document}
\ninept
\maketitle

\begin{abstract}
This work presents the task and results of the CHiME-9 challenge for Enhancing Conversations to address Hearing Impairment. The challenge considers the scenario of four-party conversations in a noisy, cafeteria-style environment with interfering speech sources and sound effects. Participants are provided with audio recordings made with Meta Aria glasses and hearing aid microphones, and clean speech samples of the conversation participants. The task is to extract the speech of the conversation partners from the noisy multi-channel recordings with the goal of improving the intelligibility and quality of the speech, evaluated using objective metrics and subjective listening tests. This paper reviews submissions from seven teams and ranks them on a combination of subjective intelligibility and quality. Results show that while the objective metrics do not reflect listener performance, the top systems were able to make substantial improvements over the challenge baseline in both intelligibility and quality ratings.
\end{abstract}

\section{Introduction}

While existing \acp{AHD} provide a substantial benefit in some scenarios, there are still many scenarios in which they face serious challenges. In particular, this issue is found in conversations in dynamic, noisy environments such as cafes, restaurants and pubs. Their limited effectiveness makes it harder for those with hearing impairments to comfortably socialise, leading to a poorer quality of life.

In recent years, research into using \acp{NN} to enhance speech for \acp{AHD} has shown great potential for improving the experience of hearing-impaired individuals. This, coupled with recent advances in low-power \ac{NN} hardware, has ushered in a new era for \acp{AHD}, with AI-enabled \acp{HA} now available to the general public.

Enhancing speech in scenarios that are both acoustically and socially complex presents a variety of challenges for \ac{NN}-based techniques. Real-world situations often contain dynamic background noise, including interfering speech, making it especially difficult for algorithms to effectively enhance the speech of a desired speaker. Further, multi-party conversations tend to have very rapid turn-taking and overlapping speech~\cite{hadley2021conversation,hadley2022timing}, where the active speaker is frequently changing, and there can also be multiple active speakers at any one time.

The CHiME-9 \ac{ECHI} challenge focuses on this scenario, providing researchers with the \ac{ECHI} dataset~\cite{sutherland2025echi} (cf.~\autoref{sec:data}), which captures real conversations in a cafeteria-style noisy environment. Teams participating in the challenge developed \ac{NN} systems to enhance speech in conversations specifically for \acp{AHD} (Aria glasses or \acp{HA}). Systems submitted to the challenge were evaluated using objective metrics, but also by a set of subjective listening tests designed to assess the speech intelligibility and quality of their systems. This paper presents the challenge task and the final results from both objective and subjective evaluation.

While the challenge is now complete, the \ac{ECHI} dataset is available for future research\footnote{\footnotesize{\url{www.huggingface.co/datasets/CHiME9-ECHI/CHiME9-ECHI}}}, and the data obtained through the listening tests will be made public in the near future.

The rest of the paper is structured as follows: a brief overview of the provided data is given in \autoref{sec:data}, the challenge task, rules and baseline are described in \autoref{sec:task}, the submitted systems are covered in \autoref{sec:subs}, the challenge results for objective and subjective assessment are presented in \autoref{sec:results} and discussed in \autoref{sec:disc} and conclusions from the challenge are presented in \autoref{sec:conc}.

\section{Data}
\label{sec:data}

Participants in the challenge were provided with the CHiME-9 \ac{ECHI} dataset~\cite{sutherland2025echi}, which contains real recordings of four-party conversations in a noisy environment. The dataset consists of $29$ hours of audio, recorded over $48$ sessions, with a total of $190$  unique participants, and contains noisy, multi-channel audio, \ac{CT} microphone audio, reference audio (for \ac{NN} training), and clean speech samples for each participant.

\subsection{Recording Scenario}

\begin{figure}[!ht]
    \centering
    \includegraphics[width=0.8\linewidth]{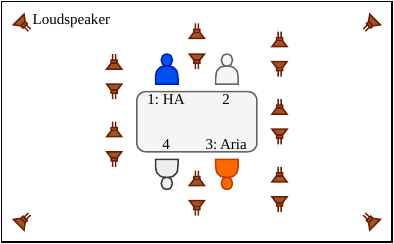}
    \caption{The recording setup for the \ac{ECHI} dataset. Note that this diagram is not drawn to scale.}
    \label{fig:echi-data}
\end{figure}

Groups of four participants would be seated around a table in the centre of the room, surrounded by $18$ loudspeakers playing background noise, as illustrated by \autoref{fig:echi-data} (not to scale). Recording sessions lasted $36$ minutes, with background noise playing at different levels throughout the session. In each session, one person would be wearing Aria glasses and another (always different from the Aria wearer) would be wearing a pair of \ac{HA} shells.

The four loudspeakers in the corners of the room played back ambient noise sourced from the WHAM! dataset~\cite{wichern2019wham}. The remaining $14$ loudspeakers surrounded the participants, and played back dynamic background noise constructed using speech from the LibriSpeech dataset~\cite{panayotov2015librispeech} and the EARS dataset~\cite{richter2024ears}, and sound effects from FSD50K~\cite{fonseca2021fsd50k}. Only sound effects which could reasonably be heard in a cafeteria were included, e.g. coughing, typing and cutlery noises.

\subsection{Materials}
\label{ssec:materials}

The noisy audio to be used as input for the algorithms is audio from Aria glasses and \ac{HA} shells. The Aria glasses record $7$-channel audio on the device, and the \acp{HA} record $4$-channel audio through an external soundcard; the \acp{HA} are worn with one on each ear, so the $4$-channel audio is made up of two channels on the left ear and two channels on the right. The Aria glasses and \acp{HA} are always worn by different participants in each session.

All conversation participants also wore a \ac{CT} microphone, which captured speech close to the mouth of the wearer at a very high \ac{SNR}, recorded through the same soundcard as the \acp{HA} (i.e.~sample-synchronous). Reference signals were generated by first applying a \ac{NN} speech denoiser~\cite{defossez2020real} to the \ac{CT} audio, as this still contained some background noise and speech from other conversation partners, followed by a delay-compensation algorithm to account for delays due to time-of-flight and clock drift~\cite{tessmer2025clockdrift} between the Aria clock and the soundcard for the \ac{CT} recordings.

Each conversation participant also recorded a reading of the first paragraph of the rainbow passage~\cite{fairbanks1960rainbow} in a quiet environment, as speech enhancement techniques can use a sample of the target speaker's voice to help extract them from a noisy mixture, e.g.~by creating speaker embeddings~\cite{cornell2023multi, hao2024xtfgrid}.

Motion tracking data was also recorded, which gives the position and orientation of each participant in the conversation. This was used in the generation of the reference signals to estimate time-of-flight, but challenge participants were also able to use it during training of their systems.

Finally, \ac{VAD} labels were provided for each conversation participant by applying a \ac{NN} \ac{VAD} algorithm to the \ac{CT} microphones~\cite{silerovad}.

\section{Challenge Task and Baseline}
\label{sec:task}

The challenge task was defined as a \ac{MSX} task, where the goal is to extract the speech from the conversation partners using the Aria/\ac{HA} recordings and rainbow passages of the conversation partners. Two separate tracks were defined for the Aria glasses and \acp{HA}; submitted systems should only process the Aria audio or \ac{HA} audio at once. Challenge participants were also expected to suppress the wearer's speech, and could use the rainbow passage of the wearer to this end.

\subsection{Data Subsets}

The dataset was split into three subsets at the session level: the \emph{training} subset contains $30$ sessions ($18$ hours), the \emph{development} subset contains $10$ sessions ($6$ hours), and the \emph{evaluation} set contains $8$ sessions ($4.8$ hours). Each subset was disjoint with respect to the conversation participants and background noise, so no speakers in the training subset would appear in the development or evaluation subsets.

For the training and development sets, all materials described in \autoref{ssec:materials} were provided to the challenge participants. For the evaluation set, only the noisy Aria and \ac{HA} audio and the rainbow passages were made public, as these were the only materials which were supposed to be used by the model.

\subsection{Challenge Rules}

The full rules to the challenge can be found on the challenge website\footnote{\footnotesize{\url{www.chimechallenge.org/challenges/chime9/task2/index}}}, but a description of the key rules is given below.

\subsubsection{Latency Requirements}

Given that the fundamental goal of the challenge is to enhance speech using \acp{AHD}, there was a strict requirement for submitted systems to operate with a $20$ ms \emph{theoretical latency}. This means that any submitted system may not use any part of the signal $>20$ ms in the future from the current time step, for example, when normalising the audio.

This latency is theoretical, as it does not account for any computation time. There is significant variation between the computational power of \acp{AHD}, so an accurate quantification of how quickly a model can run was not feasible. That said, teams were instructed to report model size and computational complexity in their reports. 

\subsubsection{External Resources}
\label{sssec:external}
A \emph{whitelist} of open-source external resources was provided to the challenge participants, ensuring that teams did not use private resources unavailable to other teams, potentially providing an advantage. Teams were able to request additions to the whitelist until $5$ months before the challenge submission deadline. 

\subsection{Baseline}

The baseline system for the challenge was a \ac{TSX} model with a TF-GridNet backbone~\cite{hao2024xtfgrid, cornell2023multi, wang2023tfgrid}. This model only extracts one speaker at a time, so it is applied iteratively with each of the target rainbow passages to extract each speaker.

\begin{figure}[!ht]
    \centering
    \includegraphics[width=0.8\linewidth]{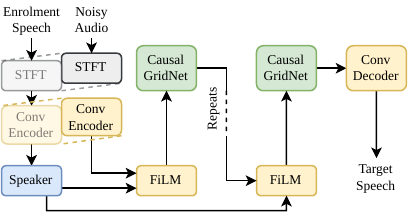}
    \caption{System diagram of the baseline \ac{TSX} network.}
    \label{fig:baseline}
\end{figure}

A system diagram for the \ac{TSX} model is given in \autoref{fig:baseline}. To ensure compliance with the latency rules, components were modified from the original TF-GridNet model~\cite{wang2023tfgrid,hao2024xtfgrid} to make them causal~\cite{cornell2023multi}. This primarily affected the \ac{STFT} and GridNet blocks, which use \ac{LSTM} networks and self-attention in the time dimension. For the \ac{STFT}, the window size was fixed to $8$ ms, with a hop size of $4$ ms. The \ac{LSTM} blocks were fixed to be unidirectional in the time dimension, and masked self-attention was used to prevent look-ahead.

\vspace{-6pt}
The baseline system was trained only using the training subset of the challenge data, with no external resources. The model hyperparameters were tuned using the \ac{STOI} scores~\cite{taal2010stoi} on the development subset. The full training parameters can be found in the challenge GitHub repository\footnote{\footnotesize{\url{www.github.com/CHiME9-ECHI/CHiME9-ECHI}}}.

\section{Submitted Systems}
\label{sec:subs}

In total, the challenge received submissions from $7$ different teams, with teams participating in the Aria track, \ac{HA} track or both; all systems are listed in \autoref{tab:results}. A wide variety of techniques was employed, with some teams extending the provided baseline and others submitting solutions based on entirely different model architectures.

\subsection{System Design}

All but one team proposed \ac{TSX} models to solve the task, with AHU-IEI~\cite{tu2026ahuiei} opting for an \ac{MSX} system instead. Eleven~\cite{zhao2026eleven} and Waseda-NTT~\cite{hu2026waseda} both introduced an \ac{OVS} block to the beginning of the system to remove the speech of the device wearer, with Eleven using the wearer's rainbow passage for \ac{NN}-based \ac{OVS}, and Waseda-NTT estimating null-beamformer coefficients. AHU-IEI only used the target speaker's rainbow passage as a training target, not using it as input to the model at all. Eleven also used the rainbow passages of the \emph{non-target} conversation partners to better disambiguate the conversation partners from one another.

\subsection{Reference Signals}

While reference signals for each conversation partner were provided in the dataset~\cite{sutherland2025echi}, the AHU-IEI and Waseda-NTT teams decided to produce their own reference signals to better remove leaked background noise/cross-talk and model transmission effects more accurately. AHU-IEI achieved this by training a speech denoiser, using data constructed from the CHiME-9 \ac{ECHI} \ac{CT} recordings, while Waseda-NTT used a close-to-distant microphone projection, which uses beamforming to estimate the transmission effects from the \ac{CT} microphone to the worn device.
\vspace{-6pt}
\subsection{Exclusion}
\label{ssec:exclusion}

During post-submission processing, one team reported generating extra training data with the CHiME-3 dataset~\cite{barker2015chime3}. Under challenge rules, this team could not be included in the final rankings as using non-whitelisted data may confer an advantage relative to compliant systems. However, the challenge organisers felt that the system was scientifically relevant in the wider context of speech enhancement for \acp{AHD}, and so retained the system in the subjective listening tests.

\section{Results}
\label{sec:results}

\begin{table*}[t]
    \centering
    \caption{The full results table for the CHiME-9 ECHI challenge. The $^\text{\textdagger}$ indicates the team excluded from the final challenge rankings (\autoref{ssec:exclusion}), and * indicates the \emph{oracle} system from the \ac{CT} microphones, with the italic objective scores indicating the slightly different objective evaluation. Bold scores indicate the best score of the systems in the official ranking. The Subjective Score is the average of the Subjective Correctness and Subjective Ovr scores, with the latter linearly mapped to the same range as the correctness. Empty rows in the subjective columns indicate teams which were not evaluated in the listening tests. Standard error is provided for the Subjective Score.}
    \label{tab:results}
    
    \resizebox{0.85\linewidth}{!}{
        \begin{tabular}{@{}cll|cccccc|ccccc@{}}
\toprule
\multicolumn{2}{c}{\multirow{2}{*}{Team}} & \multicolumn{1}{c|}{\multirow{2}{*}{System}} & \multicolumn{6}{c|}{Objective Metrics} & \multicolumn{1}{c}{Subjective} & \multicolumn{3}{c}{Subjective Quality} & \multicolumn{1}{c}{Subjective} \\
\multicolumn{2}{c}{} & \multicolumn{1}{c|}{} & \multicolumn{1}{c}{STOI} & \multicolumn{1}{c}{FW-SegSNR} & \multicolumn{1}{c}{PESQ} & \multicolumn{1}{c}{CSig} & \multicolumn{1}{c}{CBak} & \multicolumn{1}{c|}{COvr} & \multicolumn{1}{c}{Correctness} & \multicolumn{1}{c}{Sig} & \multicolumn{1}{c}{Bak} & \multicolumn{1}{c}{Ovr} & \multicolumn{1}{c}{Score} \\ \midrule
\multirow{13}{*}{Aria} & \multirow{2}{*}{ECHI} & \multicolumn{1}{l|}{Baseline} & 0.50 & 4.73 & 1.11 & 1.74 & 1.08 & 1.32 & 55.28 & 2.31 & 2.75 & 2.16 & 42.13$\pm$0.89 \\
 &  & \multicolumn{1}{l|}{CloseTalk*} & \textit{0.85} & \textit{15.41} & \textit{2.23} & \textit{3.82} & \textit{2.94} & \textit{3.05} & 86.93 & 4.57 & 3.82 & 4.31 & 84.8$\pm$0.61 \\
 & \multirow{3}{*}{AHU-IEI~\cite{tu2026ahuiei}} & \multicolumn{1}{l|}{v1-011} & \textbf{0.55} & \textbf{5.97} & \textbf{1.23} & \textbf{2.22} & \textbf{1.67} & \textbf{1.64} & \multicolumn{1}{l}{} & \multicolumn{1}{l}{\textbf{}} & \multicolumn{1}{l}{} & \multicolumn{1}{l}{\textbf{}} &  \\
 &  & \multicolumn{1}{l|}{v2-001} & 0.54 & 5.62 & 1.22 & \textbf{2.22} & 1.61 & 1.63 & \multicolumn{1}{l}{} & \multicolumn{1}{l}{} & \multicolumn{1}{l}{} & \multicolumn{1}{l}{} &  \\
 &  & \multicolumn{1}{l|}{v3-003} & 0.52 & 5.58 & \textbf{1.23} & 2.15 & 1.62 & 1.60 & 52.68 & 2.50 & 3.67 & 2.44 & 44.38$\pm$1.02 \\
 & \multicolumn{2}{l|}{ASNTU~\cite{jhou2026asntu}} & 0.50 & 5.48 & 1.18 & 2.20 & 1.56 & 1.60 & 51.86 & 1.61 & \textbf{4.11} & 1.84 & 36.41$\pm$0.90 \\
 & \multicolumn{2}{l|}{Eleven~\cite{zhao2026eleven}} & 0.45 & 0.80 & 1.08 & 1.61 & 1.06 & 1.22 & 48.34 & 3.56 & 1.81 & 2.28 & 40.17$\pm$0.95 \\
 & \multirow{2}{*}{MTEC~\cite{sharma2026mtec}} & \multicolumn{1}{l|}{Approach1} & 0.49 & 1.49 & 1.13 & 1.67 & 1.31 & 1.28 & \multicolumn{1}{l}{} & \multicolumn{1}{l}{} & \multicolumn{1}{l}{} & \multicolumn{1}{l}{} &  \\
 &  & \multicolumn{1}{l|}{Approach2} & 0.51 & 2.29 & 1.13 & 1.79 & 1.29 & 1.34 & 57.45 & \textbf{3.69} & 2.71 & \textbf{2.95} & \textbf{53.14$\pm$0.95} \\
 & \multicolumn{2}{l|}{SFU-SpeechEnhancer~\cite{haghbin2026sfu}} & 0.53 & 2.88 & 1.17 & 1.81 & 1.12 & 1.38 & \textbf{60.31} & 3.27 & 2.42 & 2.68 & 51.11$\pm$0.95 \\
 & \multirow{3}{*}{Waseda-NTT$^{\text{\textdagger}}$~\cite{hu2026waseda}} & \multicolumn{1}{l|}{RBF1\_2x} & 0.58 & 1.47 & 1.30 & 1.76 & 1.31 & 1.43 & \multicolumn{1}{l}{} & \multicolumn{1}{l}{} & \multicolumn{1}{l}{} & \multicolumn{1}{l}{} &  \\
 &  & \multicolumn{1}{l|}{RBF1\_nw\_3x} & 0.58 & 4.37 & 1.24 & 2.09 & 1.23 & 1.55 & 63.54 & 3.74 & 2.85 & 3.14 & 58.49$\pm$0.89 \\
 &  & \multicolumn{1}{l|}{RBF1\_w\_3x} & 0.57 & 2.96 & 1.25 & 2.00 & 1.22 & 1.51 & \multicolumn{1}{l}{} & \multicolumn{1}{l}{} & \multicolumn{1}{l}{} & \multicolumn{1}{l}{} &  \\ \midrule
\multirow{10}{*}{HA} & \multirow{2}{*}{ECHI} & \multicolumn{1}{l|}{Baseline} & 0.50 & 4.42 & 1.11 & 1.90 & 1.08 & 1.39 & 43.94 & 2.37 & 2.53 & 2.21 & 37.15$\pm$0.90 \\
 &  & \multicolumn{1}{l|}{CloseTalk*} & \textit{0.87} & \textit{17.77} & \textit{2.32} & \textit{3.92} & \textit{3.01} & \textit{3.15} & 85.34 & 4.54 & 3.82 & 4.26 & 83.43$\pm$0.62 \\
 & \multirow{3}{*}{AHU-IEI~\cite{tu2026ahuiei}} & \multicolumn{1}{l|}{v1-011} & \textbf{0.61} & \textbf{6.73} & \textbf{1.30} & \textbf{2.53} & \textbf{1.70} & \textbf{1.84} & \textbf{59.02} & 2.77 & 3.58 & 2.75 & \textbf{51.35$\pm$0.98} \\
 &  & \multicolumn{1}{l|}{v2-001} & 0.50 & 5.37 & 1.19 & 2.06 & 1.58 & 1.53 & \multicolumn{1}{l}{} & \multicolumn{1}{l}{} & \multicolumn{1}{l}{} & \multicolumn{1}{l}{} &  \\
 &  & \multicolumn{1}{l|}{v3-003} & 0.46 & 5.14 & 1.19 & 1.95 & 1.57 & 1.48 & 42.56 & 2.14 & \textbf{3.61} & 2.20 & 36.27$\pm$1.01 \\
 & \multicolumn{2}{l|}{ASNTU~\cite{jhou2026asntu}} & 0.47 & 5.17 & 1.13 & 2.18 & 1.47 & 1.56 & 33.72 & 1.57 & 3.57 & 1.70 & 25.57$\pm$0.86 \\
 & \multicolumn{2}{l|}{CITISIN~\cite{chao2026citisin}} & 0.53 & 5.12 & 1.17 & 2.11 & 1.39 & 1.53 & 45.24 & 2.09 & 2.95 & 2.16 & 37.10$\pm$0.95 \\
 & \multirow{2}{*}{MTEC~\cite{sharma2026mtec}} & \multicolumn{1}{l|}{Approach1} & 0.52 & 1.88 & 1.11 & 1.80 & 1.11 & 1.33 & 44.54 & \textbf{4.30} & 1.96 & \textbf{2.79} & 44.61$\pm$0.97 \\
 &  & \multicolumn{1}{l|}{Approach2} & 0.52 & 1.37 & 1.11 & 1.74 & 1.11 & 1.30 & \multicolumn{1}{l}{} & \multicolumn{1}{l}{} & \multicolumn{1}{l}{} & \multicolumn{1}{l}{} &  \\
 & \multicolumn{2}{l|}{SFU-SpeechEnhancer~\cite{haghbin2026sfu}} & 0.52 & 2.64 & 1.13 & 1.80 & 1.13 & 1.34 & 43.18 & 3.48 & 2.15 & 2.62 & 41.89$\pm$0.93 \\ \bottomrule
    \end{tabular}
}
\end{table*}

Systems were evaluated by objective measures of speech intelligibility and speech quality as well as subjective listening tests, presented in \autoref{tab:results}. The objective measures considered here are the commonly used \ac{fwSegSNR}~\cite{tribolet1978speech}, \ac{PESQ}~\cite{rix2001pesq}, and the composite speech quality metrics (CSig, CBak and COvl) ~\cite{hu2008speechmet} to assess speech quality, as well as the \ac{STOI}~\cite{taal2010stoi} to assess intelligibility. These metrics are computed with the Versa toolkit~\cite{shi2025versa}.

For subjective scoring, listening tests were conducted to assess speech intelligibility and speech quality. For each test, a listening panel of $32$ native English speakers with self-reported normal hearing was recruited. Due to the limited capacity of listening tests, teams that submitted multiple systems to the challenge were asked for their preferred system to be evaluated subjectively. These teams were provided with their objective scores on the evaluation set to inform their decision. Systems not included in the subjective evaluation have empty rows in the corresponding columns of \autoref{tab:results}.

For speech intelligibility, listeners were played a short passage ($\sim 8$ s) of the conversation and asked to transcribe the speech of a target speaker from $5$ s onwards (cued visually) under one-shot listening conditions. Systems were scored on \emph{correctness} by computing the longest common subsequence between the listener responses and the ground truth, giving percentages in the range $[0,100]$~\cite{roa2026cadenza}. The longest common subsequence rating was chosen over word error rate as it does not penalise insertions; listeners could hear multiple talkers in the audio segments, and so they should not be penalised if they write down speech from the non-target speaker. Each system received $512$ listener responses.

Speech quality was evaluated according to the ITU P.$835$ standard~\cite{itu2003835}. This involved listeners hearing $5$ s segments of audio and rating on three scales: how natural the speech sounded (Sig), how intrusive the background noise was (Bak), and overall quality defined in the context of everyday speech communication (Ovr). These were rated on continuous scales (with range $1$-$5$) with Likert-style verbal anchors. Each system received a minimum of $340$ ratings.

Included in the subjective evaluation was an \emph{oracle} system, which consisted of the summed \ac{CT} microphone channels for each session. This was included to indicate what might be achievable under near-ideal speaker extraction and to help contextualise the performance of the submitted systems. For the objective evaluation of the \ac{CT} microphones, metrics were computed using the reference signals without the delay-compensation applied, as this was only applied to account for the time-of-flight from the speaker to the device (Aria glasses or \acp{HA}), as indicated by italics in \autoref{tab:results}.

\section{Discussion}
\label{sec:disc}

All systems were able to make substantial improvements over the baseline systems in at least one of the subjective metrics, with different teams choosing to focus on different approaches to achieve the task. The Subjective Score is the average of the correctness score and the overall quality score (which was linearly mapped from its native range, $[1,5]$, to a percentage in the range $[0,100]$). 

For the \ac{HA} track, the winner by Subjective Score was AHU-IEI v1-011, with a Wilcoxon test confirming their improvement over the next best system with strong evidence ($p<0.01$). The subjective metrics suggest that this system operated primarily as a denoiser; it scored strongly on the Bak ratings, and while this caused some distortion to the naturalness of the speech, it made a huge difference to the intelligibility compared to other teams, with the next best system $13.78\%$ behind in the correctness score.

The winner of the Aria track is more tightly contested, with both MTEC-Approach2 and SFU-SpeechEnhancer performing well on the speech quality and speech intelligibility scores, respectively. A Wilcoxon test only shows weak evidence that MTEC-Approach2 outperforms SFU-SpeechEnhancer ($p<0.10$) according to the Subjective Score, so the ranking cannot be concluded only from this score. However, for the correctness ratings, SFU-SpeechEnhancer's improvements are also shown to have weak significance ($p<0.10$), while on the Ovr quality score, there is strong evidence to suggest that MTEC-Approach2 produces a greater improvement than SFU-SpeechEnhancer ($p<0.01$). With this in mind, MTEC-Approach2 is considered the winner of the Aria track.

Interestingly, MTEC-Approach2 took the converse approach to AHU-IEI-v1-011 (the \ac{HA} winners); their subjective ratings suggest that the background noise remained at an intrusive level, but the naturalness of the speech was preserved and intelligibility improved.

In general, there is very little agreement between the objective metrics and subjective results, with all objective metrics offering vastly different rankings from what is seen in the subjective results. However, it should be noted that reference signals for live recordings, such as the CHiME-9 \ac{ECHI} data, can be difficult to perfect, and so the mismatch may be related to the reference signals, as well as the metrics not quite being appropriate for conversational speech. These factors were the core of the motivation for evaluating the systems with subjective listening tests.

It can also be seen that systems typically performed better on the Aria glasses than on the \acp{HA}. This is most likely due to the inherent differences between the form factors of the devices. The Aria glasses have a comparatively fixed array geometry (subject to the arms of the glasses flexing), whereas the \acp{HA} are worn on the ears, meaning the microphones are separated by human heads, which can vary greatly in size and shape. This variation can make it more difficult for multi-channel \ac{NN}-techniques to learn inter-channel differences, leading to degraded \ac{NN} performance.

\section{Conclusions}
\label{sec:conc}

The CHiME-9 ECHI Challenge presents the task of enhancing the speech of conversation partners in noisy environments using recordings of real conversations in a simulated cafeteria-style environment. The challenge also evaluates submissions using both objective metrics and subjective listening tests of speech intelligibility and speech quality. Submissions varied greatly in how they approached the task, both from a technical and methodological approach, yielding improvements in all metrics over the challenge baseline, with the top systems making substantial gains in both the intelligibility and quality of the speech. These results are very encouraging for research into improving the experience of hearing-impaired listeners in social situations, and will hopefully promote future work in this field.

\bibliographystyle{IEEEbib}
\bibliography{refs}

\end{document}